\documentclass[letterpaper, 10 pt, conference]{ieeeconf}  % Comment this line out if you need a4paper

\IEEEoverridecommandlockouts                              % This command is only needed if
\usepackage{graphicx} % for pdf, bitmapped graphics files
\graphicspath{{./fig/}}
\DeclareGraphicsExtensions{.pdf}
\usepackage{cite}
\usepackage{amsmath} % assumes amsmath package installed
\usepackage{color}
\usepackage{algorithm}
\usepackage[noend]{algpseudocode}
\makeatletter
\renewcommand{\ALG@beginalgorithmic}{\small}
\makeatother

\DeclareMathOperator{\arctantwo}{arctan2}

\usepackage{caption}

\usepackage{etoolbox}
\apptocmd{\sloppy}{\hbadness 10000\relax}{}{}

\renewcommand{\baselinestretch}{0.9815}

\title{\LARGE \bf
Ex-DoF: Expansion of Action Degree-of-Freedom\\with Virtual Camera Rotation for Omnidirectional Image%
}

\author{Kosuke Tahara$^{1}$ and Noriaki Hirose$^{1}$% <-this % stops a space
\thanks{$^{1}$Kosuke Tahara and Noriaki Hirose are with Toyota Central R\&D Labs., Inc. {\tt\small ktaha@mosk.tytlabs.co.jp}}%
}

\begin{document}

\maketitle
\thispagestyle{empty}
\pagestyle{empty}

\begin{abstract}

Inter-robot transfer of training data is a little explored topic in learning- and vision-based robot control. Here we propose a transfer method from a robot with a lower Degree-of-Freedom (DoF) to one with a higher DoF utilizing the omnidirectional camera image. The virtual rotation of the robot camera enables data augmentation in this transfer learning process. As an experimental demonstration, a vision-based control policy for a 6-DoF robot is trained using a dataset collected by a wheeled ground robot with only three DoFs. Towards the application of robotic manipulations, we also demonstrate a control system of a 6-DoF arm robot using multiple policies with different fields of view to enable object reaching tasks.

\end{abstract}

\section{Introduction}
Vision-based robot control has been enabled by recent advances in machine learning and image recognition\cite{taiSurveyDeepNetwork2018}.
While depth information is utilized in mainstream robot control~\cite{thrun2002probabilistic,biswas2012depth},
LiDAR or RGB-D cameras can be expensive and prone to failures that are related to light reflection or interference\cite{martinmartinDeteriorationDepthMeasurements2014,tangPerformanceTestAutonomous2020}.
Conversely, the RGB camera operates in a wider range of environmental conditions and its hardware cost is relatively low.
In addition, it can provide a wider field of view (FoV) easily; an extremal of that is omnidirectional vision ($360^\circ$\,FoV).
The $360^\circ$\,FoV without intrinsic blind spots can provide robust and safe control policies\cite{hirose2018gonet,hirose2019deep}.

When machine learning is employed in the processing of visual information, generalization performance is a fundamental problem\cite{zhouDomainGeneralizationVision2021}.
Given that deep-learning-based image processing requires a vast amount of training data for effective generalization,
it is necessary to collect robotic vision data in a wide variety of environments\cite{sunderhaufLimitsPotentialsDeep2018}.
Data collection is relatively acceptable for mobile robots~\cite{hirose2018gonet,Burri25012016,Geiger2012CVPR,Cordts2016Cityscapes}, however, its cost can be problematic for robots with low mobility, such as robotic arms\cite{DBLP:conf/corl/SharmaMPG18,DBLP:conf/corl/DasariETNBSSLF19,DBLP:conf/iros/MandlekarBSTGZG19,ebertBridgeDataBoosting2021}.
If the data are dependent on the robot type, this issue will be much more severe;
data must be collected for each robot in different environments.

\begin{figure}
  \centering
  \includegraphics[width=0.85\hsize]{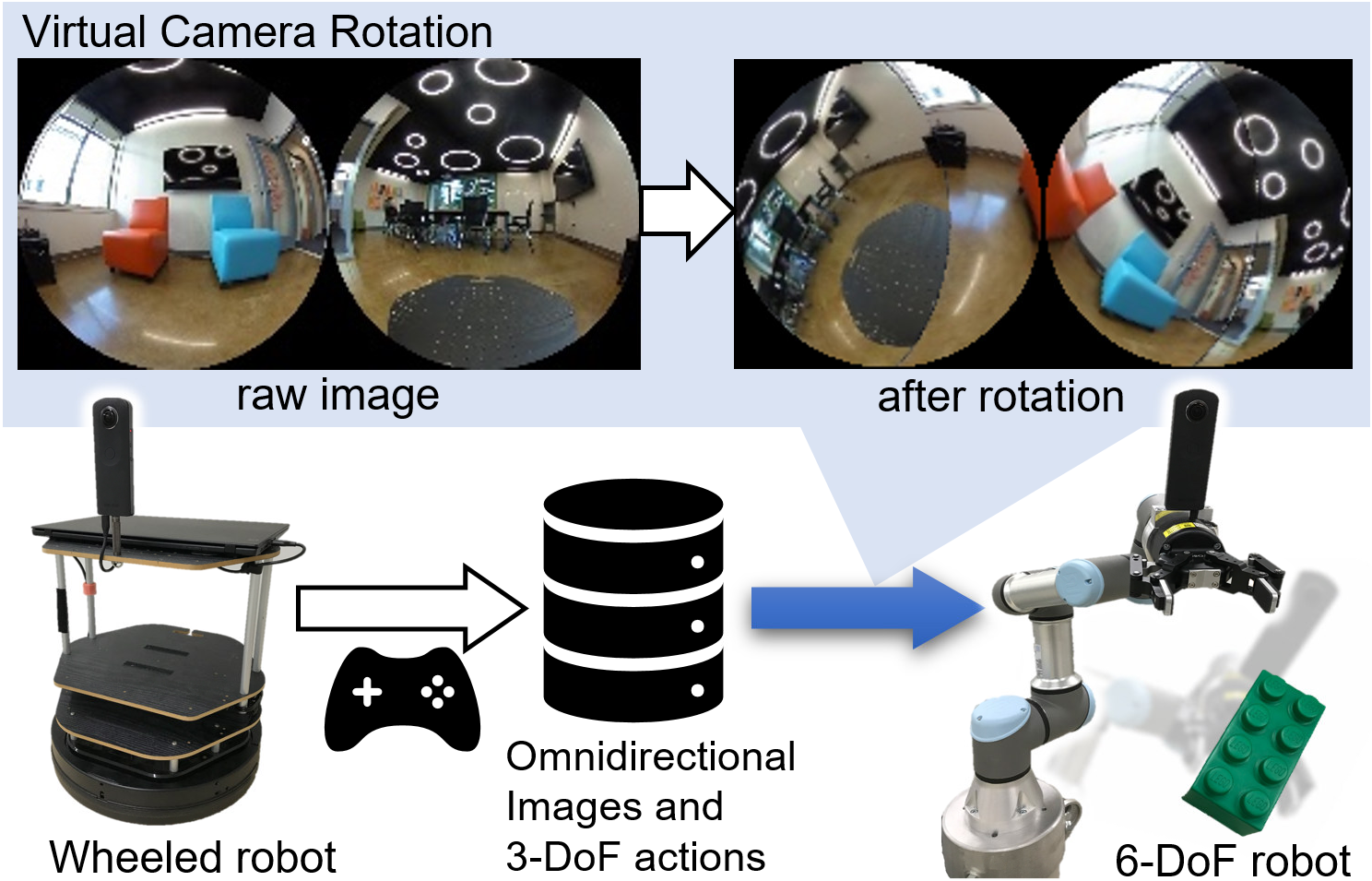}
  \caption{\small {\bf Overview of our method.}
  The training data is acquired by a wheeled robot traveling on flat floors (3-DoF).
  The virtual camera rotation makes it possible to train a control policy of 6-DoF action for the robotic arm using this dataset only.
  } \label{fig:concept}
  \vspace{-6mm}
\end{figure}

This paper proposes a method to deal with this problem through the inter-robot transfer of training data on the vision-based robot control.
After leveraging properties of omnidirectional vision, the Degree-of-Freedom (DoF) of trainable action is expanded.
To be concrete, a deep neural network (DNN) is trained for the control of a 6-DoF robot
using a dataset taken only by a 3-DoF differential wheeled ground robot.
This concept is illustrated in Fig.~\ref{fig:concept}.
Since the camera image contains omnidirectional information, the camera can be rotated virtually in an arbitrary direction.
Data augmentation using virtual camera rotation (VCR) expands the DoF of the actions contained in the training dataset.
Our DNN model takes current and target images as inputs, and outputs actions to match the current vision with the target.
Since our control policy generates 6-DoF actions based on the camera coordinate,
it will apply to various robots equipped with the same camera.
In this paper, we show an application to a 6-DoF robotic arm.

Given that the policy is based on omnidirectional images, it is robust to small changes in the environment
because such a change occupies a small area in the entire image.
In other words, it can fail to capture local information in the environment.
This property makes it difficult to apply the policy to realistic problems of object reaching and grasping,
e.g., when the object is placed at different local positions on target-acquisition and task-execution.

The second novelty of this study is a control system for manipulation tasks that resolves the above problem.
In the proposed system, another policy with a limited FoV is trained for local control, and the robot is controlled using combined two policies,
i.e., omnidirectional (global) and limited (local).
The proposed method is quantitatively and qualitatively evaluated with a robot simulator~\cite{shenIGibsonSimulationEnvironment2021} and a real robotic arm.

\section{Related Work}
\label{sec:related}
%\subsection{Model-Based Visual Servoing}
%\label{sec:related-model}
\noindent
{\bf Model-Based Visual Servoing: }
Visual servoing has been tackling a similar problem of minimizing the difference between the target and current visions\cite{hutchinsonTutorialVisualServo1996,chaumetteVisualServoControl2006,chaumetteVisualServoControl2007}.
The traditional approach of visual servoing is model-based, where the image Jacobian is defined to relate the robot action to image-space.
The main limitations of this approach are robustness against changes in the environment and the requirement of the image Jacobian.
Visual model predictive control (MPC) studies have formulated MPC problems for visual servoing\cite{sauveeImageBasedVisual2006,liVisionBasedModelPredictive2016,calliVisionbasedModelPredictive2017}.
These methods also involve explicit modeling of robot kinematics and/or dynamics.
The model-based nature makes it almost impossible to transfer across different robots or environments.

%\subsection{Learning-Based Visual Control}
%\label{sec:related-learning}
\noindent
{\bf Learning-Based Visual Control: }
Recent advances in machine learning have driven investigations into learning-based visual control for robots.
Deep reinforcement learning (RL) methods encode visual observation into a compact state representation using DNNs to perform RL\cite{zhuTargetdrivenVisualNavigation2017,kalashnikovScalableDeepReinforcement2018}.
While the agent can learn complex tasks, it requires a well-designed reward and many train iterations for sufficient exploration.
They especially face difficulties during training in real-world environments\cite{dulac-arnoldChallengesRealworldReinforcement2021}.

On imitation learning (IL), the control policy is trained using human demonstrations\cite{argallSurveyRobotLearning2009,codevillaE2E2018,pathakICLR18zeroshot,xuNeuralTaskProgramming2018,mandlekarGTILearningGeneralize2020}.
This approach provides a safe and efficient way to learn complex skills; however,
it is difficult to generalize towards the state space outside the collected demonstration.
Furthermore, the data collection by human demonstration is usually expensive.

Another approach is to train a visual predictive model conditioned by the robot action\cite{finn2017,hirose2018gonet,hiroseVUNetDynamicScene2019,hirose2019deep,hiroseProbabilisticVisualNavigation2020}.
Finn \textit{et al}. \cite{finn2017} used a trained predictive model for a robotic arm to perform the stochastic optimization during execution.
Hirose \textit{et al}. \cite{hirose2019deep,hiroseProbabilisticVisualNavigation2020} trained another DNN for action inference through a predictive model to enable long-horizon navigation.
Although DNN-based predictive models have been proven to enable generalizable visual control,
abundant collected data are used only for a single type of robot.

%\subsection{Generalization}
%\label{sec:related-generalize}
\noindent
{\bf Generalization: }
Previous studies have tackled the problem of generalization.
Several frameworks were proposed to achieve generalizable IL policies in terms of executable tasks and domains\cite{xuNeuralTaskProgramming2018,mandlekarGTILearningGeneralize2020,ebertBridgeDataBoosting2021}.
The inter-robot transfer of RL policy has also been investigated\cite{guptaLearningInvariantFeature2017,devinLearningModularNeural2017,chenHardwareConditionedPolicies2018,kangHierarchicallyIntegratedModels2021}.
Since our approach is a transferred training of visual predictive model instead of IL or RL,
the most related works are \cite{DBLP:conf/corl/DasariETNBSSLF19} and \cite{DBLP:conf/eccv/SchmeckpeperXRT20},
which demonstrate the training of predictive models across multiple embodiments.

However, the aforementioned works did not aim for the generalization to another action space with more dimensions.
A notable feature of our work is that the transfer is demonstrated between embodiments with a large DoF gap: wheeled (3-DoF) and arm (6-DoF) robots.
In addition, our method is proved to work well on a real robot as well as in the simulator.
\section{Data Augmentation for Transfer}
\label{sec:method}
Our method aims to acquire a robot control policy that outputs 6-DoF robot actions $\mathbf{a}=\left(x, y, z, \alpha, \beta, \gamma \right)$ from the inputs of the current omnidirectional camera image $I(0)$ and the target image $I_G$.
Here, the action is defined as a relative pose of the camera.
$\left(x, y, z\right)$ denotes 3D translation and $\left(\alpha, \beta, \gamma\right)$ is 3D rotation represented as roll, pitch, and yaw angles.
The critical point of our problem is that the training dataset is collected solely by a wheeled robot.
Given that the wheeled robot travels only on flat floors, the action has only three DoFs, i.e.,
the components $z$, $\alpha$, and $\beta$ are always zero.

Our data augmentation method expands DoF to overcome this barrier by virtually rotating the camera pose, leveraging a property of the omnidirectional camera.
During policy training, data pairs from the 3-DoF dataset $\left( I_d(i), \mathbf{a}_d(i) \right)$ can be converted to 6-DoF training data pairs $\left( I(i), \mathbf{a}(i) \right)$ through the VCR.
A random 3D rotation is generated per data sample for the VCR, and we denote its homogeneous transformation matrix $T$.
As described below, the VCR is purely geometric and deterministic transformation.

\subsection{Virtual Camera Rotation (VCR) $I_d(i) \to I(i)$}
\label{sec:method-rot}
In this study, we assume that the omnidirectional camera is a dual-fisheye type camera,
and its image consists of front and back circular fisheye images.
For images in Figs. \ref{fig:concept}, \ref{fig:ctrl-sys}, \ref{fig:fisheye-rot}, \ref{fig:fine-mask}, and \ref{fig:visual}, the front (back) image is on the left (right).
The VCR converts a dual-fisheye image to another one as if it is taken after the camera is rotated arbitrarily in three dimensions.

It is assumed that the optical system of the fisheye camera has rotational symmetry around the optical axis.
The incident light from the $\left(\theta, \phi\right)$ direction in 3D polar coordinates of real space is projected to $\left(r(\theta), \phi\right)$ in 2D polar coordinates of the image-space~\cite{courbonGenericFisheyeCamera2007,hughes2010accuracy,kumar2020fisheyedistancenet}.
The $r(\theta)$ is called the projection function, which depends on camera design and manufacturing variation.
Calibration is necessary to identify $r(\theta)$ precisely; however,
it is also valuable to test the method using simple and generic projection functions
in order to show the applicability to uncalibrated cameras.
For this reason, we assume an equidistant projection function $r(\theta)\propto\theta$.

\begin{algorithm}[b]
  \caption{Appearance Flow by Camera Rotation $T^{\mu\nu}$}
  \label{alg:rot}
  \begin{algorithmic}[1]
    \Function {BackProj}{$u, v$}
        \State {$\theta \gets r^{-1} \left(\sqrt{u^2+v^2}\right) = \pi \sqrt{u^2+v^2} $} \label{alg:rot-rinv}
        \State {$\phi \gets \arctantwo{(v, u)}$}
        \State {\Return {$(x, y, z) = (\sin{\theta}\cos{\phi}, \sin{\theta}\sin{\phi}, \cos{\theta}$)}}
    \EndFunction
    \Function {Proj}{$x, y, z$}
        \State {$R \gets r \left(\arccos{(z)}\right) = \arccos{(z)} / \pi $}
        \label{alg:rot-r}
        \State {$\phi \gets \arctantwo{(y, x)}$}
        \State {\Return {$(u, v) = (R\cos{\phi}, R\sin{\phi})$}}
    \EndFunction
    \Function {Flow}{$u, v, T^{\mu\nu}$}  \Comment{main computation}
    \State {$Q_{hs} \gets$ \Call {BackProj} {$u, v$}}
    \State {$Q^{\mu\nu}_T \gets$ $T^{\mu\nu}Q_{hs}$}
    \State {\Return $M^{\mu\nu}_T$ = \Call {Proj} {$Q^{\mu\nu}_T$}} \label{alg:rot-Tp}
    \EndFunction
  \end{algorithmic}
\end{algorithm}

\begin{figure}[t!]
  \vspace{2mm}
  \centering
  \includegraphics[width=8.5cm]{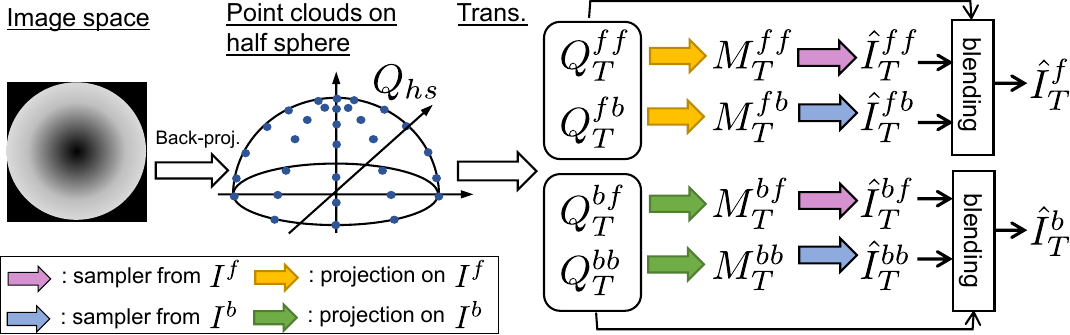}
  \caption{\small {\bf Procedure of VCR.}
  Each pixel in the image is back-projected to the point on unit half-sphere in real space.
  The point cloud is rotated to $Q^{\mu\nu}_T$ and re-projected to the image plane, constructing flow $M^{\mu\nu}_T$.
  The original front and back images are sampled and blended to predict rotated images.
  }
  \vspace{-6mm}
  \label{fig:pose-rot}
\end{figure}

The VCR is performed by sampling the original image using the appearance flow\cite{zhouViewSynthesisAppearance2016}.
Since the involved transformation is rotation only,
the flow can be computed geometrically, as outlined in Algorithm~\ref{alg:rot} and visualized in Fig.~\ref{fig:pose-rot}.
A standardized image-space point $(u, v)\in[-1,1]^2$ is first back-projected to the real space point $(x, y, z)$ on unit half-sphere
\footnote{This method is inspired by self-supervised monocular depth estimation\cite{zhouUnsupervisedLearningDepth2017,godard2019digging,vijayanarasimhanSfMNetLearningStructure2017}. However, depth information is not required and we can assume uniform depth (point clouds on unit half-sphere) because only rotational transformation is considered.},
transformed by $T^{\mu\nu}$, and projected again to the image-space.
It is noteworthy that the right-hand sides of lines \ref{alg:rot-rinv} and \ref{alg:rot-r} are the forms for equidistant projection with an FoV of $\pi$ radians; however, this algorithm is applicable to any invertible projection function $r$.

For the dual-fisheye image, some pixels in the front image correspond to pixels in the back image after rotation, and \textit{vice versa}.
This interchange can be resolved by generating multiple predicted images and blending them, as shown in Fig.~\ref{fig:pose-rot}.
Given the camera rotation $T$, the transformations between camera frames $T^{\mu\nu}$ are computed for four different patterns of $\mu\nu \in \{ ff, fb, bf, bb\}$, where e.g. $fb$ is read as ``\textit{back} to \textit{front} through rotation''.
Assuming $T$ is given on the front camera frame and $T_l$ converts front to back camera frames, and \textit{vice versa}, these transformations are written as
\begin{equation}
    T^{ff} = T,\hspace{1mm}T^{fb} = T_l T,\hspace{1mm}T^{bf} = T T_l,\hspace{1mm}T^{bb} = T_l T T_l. \label{eq:Tff}
\end{equation}
The corresponding point clouds $Q^{\mu\nu}_T$ and flows $M^{\mu\nu}_T$ are computed by Algorithm~\ref{alg:rot}.
The image predictions are performed using sampling function $f_{s}$ as $\hat{I}^{\mu\nu}_T = f_{s} (M^{\mu\nu}_T, I_{\nu} )$ \cite{jaderbergSpatialTransformerNetworks2015}.
Finally, four images are blended to predict front and back images as $\hat{I}^{\mu}_T = \hat{I}^{\mu f} [ Q_T^{\mu f}(z) \geq 0] + \hat{I}^{\mu b} [ Q_T^{\mu b}(z) \geq 0 ]$.
After assuming an FoV of $\pi$, we used the $z$ coordinate of point clouds $Q^{\mu\nu}_T(z)$ for filtering here,
in order to determine whether a pixel should actually be incorporated in the prediction.

\subsection{Action DoF Expansion $\mathbf{a}_d(i) \to \mathbf{a}(i)$}
\label{sec:method-dof}
\begin{figure}
  \vspace{2mm}
  \centering
  \includegraphics[width=7.5cm]{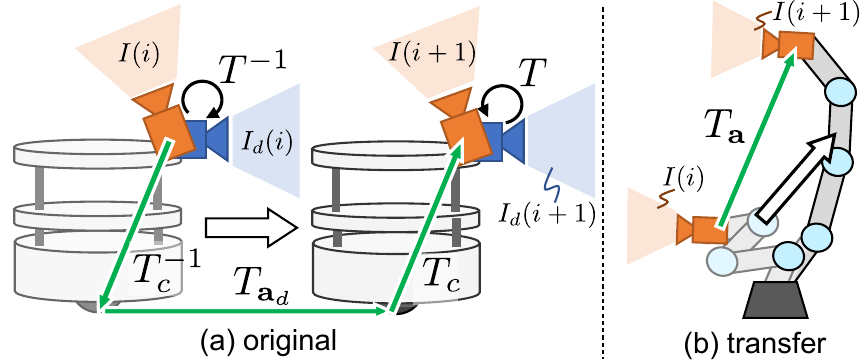}
  \caption{\small {\bf Transformation matrix $T_{\mathbf{a}}$.}
  (a) Wheeled robot and camera heading right on data acquisition.
  $T$ changes camera direction virtually to left-up direction and
  $T_{\mathbf{a}}$ (transformation of the left-up directed camera from past to future)
  contains a going-up (non-zero $z$) component, which never appears in the original data.
  (b) After transfer, 6-DoF control of the robotic arm is enabled by the DoF expansion.
  }
  \label{fig:pose-trans}
  \vspace{-6mm}
\end{figure}

As shown in Fig.~\ref{fig:pose-trans}, a transformation matrix of 6-DoF action $T_{\mathbf{a}}(i)$ can be computed as $T_{\mathbf{a}}(i) = T^{-1} T^{-1}_{c} T_{\mathbf{a}_d}(i) T_{c} T$.
Here, $T_c$ is a fixed transformation from the robot base to the camera in the setup for data acquisition.
$T_{\mathbf{a}_d}(i)$ is a transformation matrix corresponding to the original 3-DoF action $\mathbf{a}_d(i)$ represented on the robot base frame.
Since the wheeled robot travels in 2D space,
the components of $T_{\mathbf{a}_d}(i)$ corresponding to $z$, $\alpha$, and $\beta$ are always zero.
However, $T_{\mathbf{a}}(i)$ can contain non-zero $z$ translation and roll/pitch rotations.
This data augmentation makes it possible to train policy for 6-DoF action using a dataset acquired by a 3-DoF robot only.

We believe that giving random 3D rotations $T$ is important for the DoF expansion because it removes the model's dependency on the absolute camera pose in the environment.
For example, the floor is a special instance for the wheeled robot in building environments since the camera is always at a constant height from the floor.
However, this is not the case after random VCR,
because the floor's pose in the image is also randomized and it becomes indistinguishable from other flat things such as walls or ceilings.
After a VCR depicted in Fig.~\ref{fig:pose-trans}, for instance, augmented action is a going-up motion with a virtual wall (floor) behind.

\section{Control System}
\label{sec:ctrl}
In this section, we describe the control system for a 6-DoF robotic arm with policies trained using VCR.
An omnidirectional camera is mounted on the robot's end-effector to control its position.
Inspired by \cite{hirose2019deep}, we first introduce an MPC-based approach to train ``EnvNet'',
which controls the end-effector in order to match the vision with a target.
Then, we propose a control system utilizing two different policy models, ``EnvNet'' and ``ObjNet'' to enable object reaching.

\subsection{MPC Based Training with VCR}
Our control policy is based on MPC and attempts to minimize a cost function defined as the difference between target image $I_G$ and predicted images $\hat{I}(i)$ on the prediction horizon $\mathcal{T}=\{1,\ldots,N\}$~\cite{hirose2019deep}.
By minimizing the image difference in training, the trained policy can move the robot to match the current vision $I(0)$ with $I_G$.

\begin{figure}
  \vspace{2mm}
  \centering
  \includegraphics[width=7.8cm]{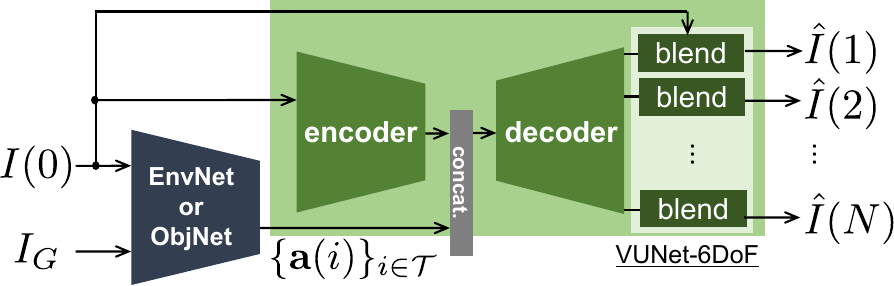}
  \caption{\small {\bf Neural Network models.}
  The shadowed (green) area indicates VUNet-6DoF, which predicts the future camera image based on the current image and actions.
  EnvNet and ObjNet generate actions from current and target images.}
  \label{fig:vunet-polinet}
  \vspace{-6mm}
\end{figure}

First, only VUNet-6DoF (shadowed part in Fig.~\ref{fig:vunet-polinet}) is trained to predict $N$ images conditioned on virtual $N$ step 6-DoF actions $\{\mathbf{a}(i)\}_{i\in\mathcal{T}}$, through minimizing mean absolute pixel difference $J_v$:
\begin{equation}
  J_{v} = \frac{1}{NN_{pix}}\sum_{i=1}^{N}{
  \left|I(i)-\hat{I}(i)\right|
  }, \label{eq:Jvunet}
\end{equation}
where $N_{pix}$ is the total number of pixels in an image.
Note that $\{I(i)\}_{i\in\{0\}\cup\mathcal{T}}$ and $\{\mathbf{a}(i)\}_{i\in\mathcal{T}}$ are converted by VCR to incorporate the 6-DoF action.
After training VUNet-6DoF, we train the control policy, EnvNet, through fixed VUNet-6DoF.
The loss function for EnvNet training is defined as the sum of two terms $J_{p} = J_{i} + k_r J_{r}$, where
\begin{align}
  J_{i} &= \frac{1}{NN_{pix}}\sum_{i=1}^{N}{\left|
    I_G-\hat{I}(i)
    \right|}, \label{eq:Ji} \\
  J_{r} &= \frac{1}{6N} \sum_{i=1}^{N} \left(
    \left(x_i^2 + y_i^2 + z_i^2\right) + \left(\alpha_i^2 + \beta_i^2 + \gamma_i^2\right)
    \right).
\end{align}
$J_i$ is the difference between the target image and VUNet-6DoF prediction when actions are performed.
$J_r$ is a regularization factor that suppresses the norm of actions,
and the parameter $k_r$ determines its strength.
The images are also augmented here by the VCR.
The trained EnvNet can generate 6-DoF virtual actions $\mathbf{a}(i)$ for vision-matching.
\subsection{Combined Control System for Object Reaching Task}
Since the entire FoV is considered, the EnvNet can robustly control the end-effector toward the target position even when there are small changes in the environment.
However, the robotic arm control should focus on the local object in manipulation tasks.
If the position of the target object is different between target-acquisition time and task-execution time,
EnvNet may regard the difference as a ``small change'' in the environment and cannot control the robot toward the object.
This is fatal in grasping tasks, for instance.

\begin{figure}
  \vspace{2mm}
  \centering
  \includegraphics[width=8.0cm]{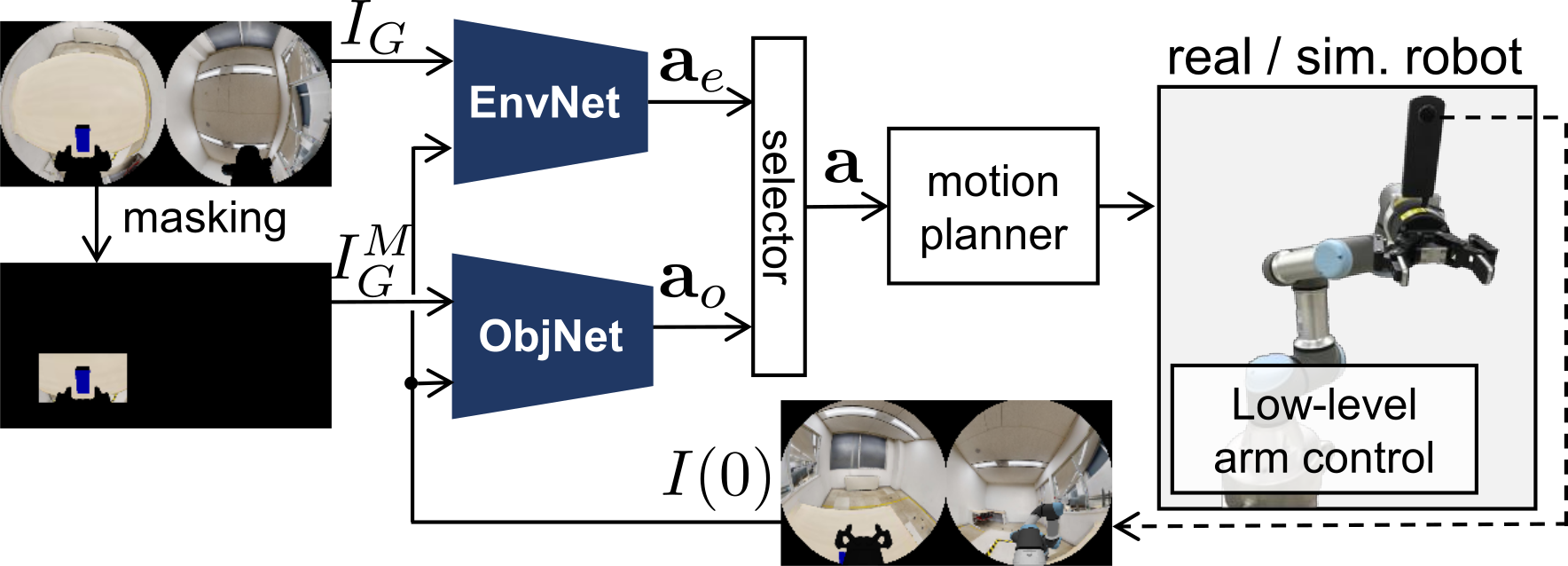}
  \caption{\small {\bf The control system for the robotic arm using two policy networks.}
  EnvNet and ObjNet output individual actions by using different FoV.
  The selector picks the suitable action and passes it to the motion planner.
  The motion planner generates joint trajectory command for the low-level controller.
  } \label{fig:ctrl-sys}
  \vspace{-3mm}
\end{figure}

To overcome this problem, we train another policy named ObjNet and propose a combined system with EnvNet, as shown in Fig.~\ref{fig:ctrl-sys}.
The ObjNet is trained to consider a fixed Region of Interest (RoI) in the image.
During the training of ObjNet, the region outside the RoI in $I_G$ is masked out, and the loss function is calculated only inside the RoI.
We can predetermine the RoI because the target object is commonly in front of the camera (and the end-effector) on grasping
and $I_G$ will be captured at this grasping posture.

During execution, the current and target images are fed into EnvNet and ObjNet to obtain action candidates $\{\mathbf{a}_e(i)\}$ and $\{\mathbf{a}_o(i)\}$ $\forall i\in \mathcal{T}$.
The ``selector'' in Fig.~\ref{fig:ctrl-sys} selects the action $\mathbf{a}$ to execute from the $2N$ candidates.
Since the action is expressed as a relative camera pose,
a motion planner is used to plan the robot actuation command.
The motion planner computes the joint-space trajectory from the current robot pose and requested action $\mathbf{a}$,
and sends it to a low-level arm controller in the real robot or the simulator.

Further, there is some room for the design of the selection algorithm of $\{\mathbf{a}_e(i)\}$ and $\{\mathbf{a}_o(i)\}$.
For simplicity, we have chosen to evaluate the absolute values of each element of $\{\mathbf{a}_e(i)\}$,
and use $\{\mathbf{a}_o(i)\}$ only if all the elements are smaller than the predetermined thresholds: $\eta_p$ and $\eta_r$ for position and rotation elements, respectively.
This criterion assures activation of EnvNet for a larger movement at first and subsequent fine-tuning by ObjNet.

Once an action sequence $\{\mathbf{a}(i)\}$ is selected from the two, step $i$ is chosen.
As this method is based on receding horizon control, the standard choice is always the nearest future, i.e., $i=1$.
However, the motion planner can fail for some actions,
because the policies do not consider the arm pose and action is predicted by the camera images only.
In the case of failure, the motion planner is sequentially fed from the next-nearest actions,
and the first action with successful planning is executed.
If plannings for all the actions $\{\mathbf{a}(i)\}_{i\in\mathcal{T}}$ have failed, the system is considered to be stuck.
\section{Experiments}
\label{sec:expt}
\subsection{Datasets}
\label{sec:expt-dataset}
Two datasets were used for the policy training.
One is the Go Stanford 4 (``GS'') dataset\cite{gs4dataset} to evaluate our methods.
The dataset was created using a wheeled robot (Turtlebot2) equipped with a dual-fisheye camera (Ricoh Theta S) and
contains pairs of teleoperation commands (translational and angular velocities) and camera images.
The camera images were prepared by clipping the raw image to obtain an FoV around $\pi$.
In the evaluation, ``GS$^+$'' denotes the GS dataset augmented by VCR.
``GS$^-$'' indicates the raw GS dataset.

Another dataset named ``Arm'' was collected by simulated arm robot UR3e with an omnidirectional camera at the end-effector in the iGibson \cite{shenIGibsonSimulationEnvironment2021} version 1.0.3.
The simulated camera image and pose data were collected while the robot was sequentially moving towards randomly sampled 6D poses.
This simulated robot was also used for quantitative evaluation.
The data were collected in three simulated environments,
and the total number of training data $>2\times10^5$ is similar to that of the GS dataset.

\subsection{Data Augmentation}
\label{sec:expt-aug}
\begin{figure}
  %\vspace{1mm}
  \centering
  \includegraphics[width=8.5cm]{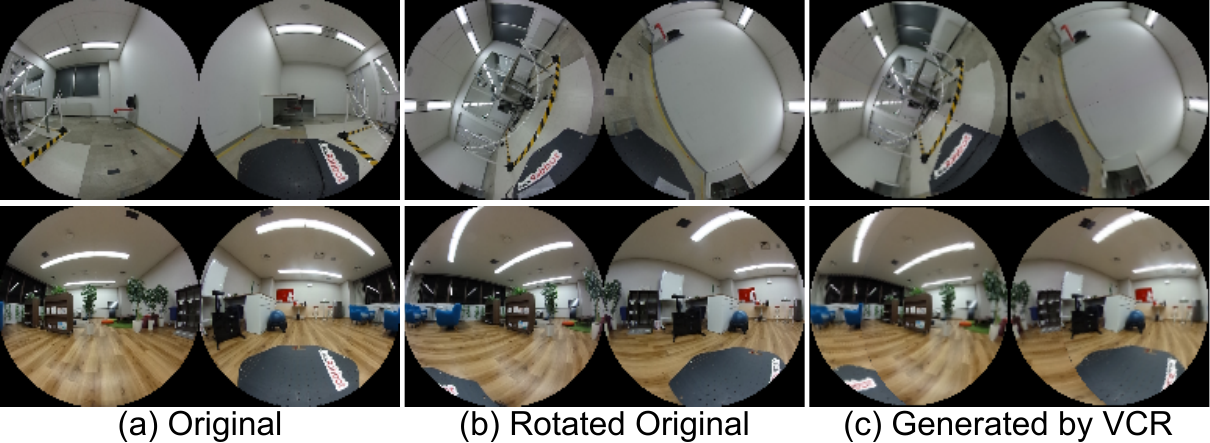}
  \caption{\small {\bf Visualization of VCR.}
  (a) Original images without camera rotation.
  (b) Images with camera rotated by $(\alpha, \gamma)=(68^\circ, 88^\circ)$ (top) $(0^\circ, 45^\circ)$ (bottom).
  (c) Generated images from (a) by VCR.
  } \label{fig:fisheye-rot}
  \vspace{-3mm}
\end{figure}

We qualitatively validated the VCR method.
Fig.~\ref{fig:fisheye-rot} (a) shows dual-fisheye images taken by Theta S mounted on Turtlebot2.
Images in Fig.~\ref{fig:fisheye-rot} (b) are taken after actually rotating the camera poses in the roll and yaw directions by $(\alpha, \gamma)=(68^\circ, 88^\circ)$ (top) and $(0^\circ, 45^\circ)$ (bottom).
The generated images by the corresponding VCR are in Fig.~\ref{fig:fisheye-rot} (c).
We can confirm that Fig.~\ref{fig:fisheye-rot} (c) is similar to (b) and realistic,
although there are some imperfections due to the use of uncalibrated projection function.
It is assumed that small discrepancies in images will not result in notable negative effects on the training of the control policy in this work.

\subsection{Training}
\label{sec:expt-train}
\subsubsection{Network Structure}
Here, we explain the network structures shown in Fig.~\ref{fig:vunet-polinet}.
VUNet-6DoF is used for the prediction of future images on $N=8$ steps.
The current image $I(0)$ ($128\times128$ pixels per front and back fisheye images, RGB channels) is encoded to a feature vector of 512 dimensions.
The encoder is a standard convolutional neural network (CNN) with eight layers, and a layer consists of convolution, batch normalization, and ReLU activation.
The encoded feature is concatenated with actions $\{\mathbf{a}(i)\}_{i\in\mathcal{T}}$ and fed into the decoder to obtain the appearance flow ($2\times128\times128$) and visibility mask ($1\times128\times128$) for image prediction~\cite{jaderbergSpatialTransformerNetworks2015} at each step $i$.
%The appearance flow maps the pixel location of current image to predicted ones.
%Blending module samples current image using the flow to generate predicted front and back fisheye images,
%and two images are further blended using the visibility mask to output final prediction $\hat{I}(i)$.
The studies in \cite{hirose2019deep} and \cite{hiroseVUNetDynamicScene2019} provide further details on the VUNet concept and structure.

The control policy networks EnvNet and ObjNet output the robot actions $\{\mathbf{a}(i)\}$ based on the current image $I(0)$ and given target image $I_G$.
These two networks are CNNs with a structure similar to the encoder part of VUNet-6DoF.

\subsubsection{Training Procedures}
First, VUNet-6DoF was trained by minimizing $J_v$ using the Adam optimizer\cite{kingmaAdamMethodStochastic2015} with a learning rate of 0.001.
A subsequence of images and actions $\left( I_d(0), I_d(i), \mathbf{a}_d(i) \right)$ $(\forall i \in \mathcal{T})$ was sampled from the dataset and converted into training data $\left( I(0), I(i), \mathbf{a}(i) \right)$ using the technique described in Sec. \ref{sec:method}.
After that, policy networks were trained using pairs of current and target images $\left( I(0), I_G \right)$,
which were converted by random VCR from the original image pairs $\left( I_d(0), I_d(i) \right)$,
where $i$ is uniformly sampled from $\mathcal{T}$.
We set $k_r=0.1$ for $J_p$.

\begin{figure}
  \vspace{2mm}
  \centering
  \includegraphics[width=7.5cm]{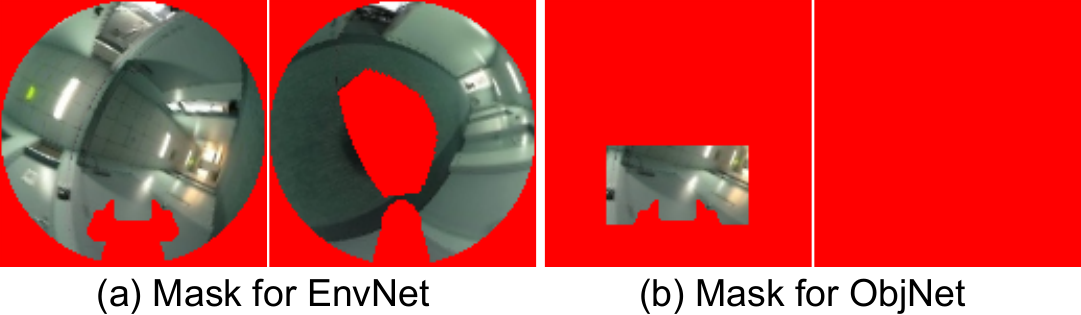}
  \caption{\small {\bf Masks for policy networks training.}
  (a) An augmented image from GS dataset with mask for EnvNet training.
  (b) with mask for ObjNet training.
  Filled (red) areas are masked regions.
  } \label{fig:fine-mask}
  \vspace{-5mm}
\end{figure}

As described in Sec.~\ref{sec:ctrl}, ObjNet was trained so that the policy focuses on a specified RoI in the image.
Thus, the area outside the RoI was masked out in the target image $I_G$ and predicted image $\hat{I}(i)$ in the loss function computation of Eq.~(\ref{eq:Ji}).
Fig.~\ref{fig:fine-mask} shows an example of the masked images.
It is noteworthy that even for EnvNet training (Fig.~\ref{fig:fine-mask} (a)), several regions were masked out: where the arm robot itself will be shown, the wheeled robot in GS dataset is shown, and outside fisheye image circle.
These masks were employed to avoid the prediction disturbance by static objects
fixed at camera's frame on training (wheeled) and runtime (arm robot).
For the arm robot, we masked out the parts near the end-effector, which are always in the image regardless of the robot pose.
As shown in Fig.~\ref{fig:fine-mask} (b), only a rectangular RoI around the gripper in the front image was used for ObjNet training,
where the object will be placed on the object reaching task.
The back-side image was completely masked out.

\subsection{Evaluation Conditions}
\label{sec:expt-cond}
\begin{figure}
  \vspace{2mm}
  \centering
  \includegraphics[width=8.5cm]{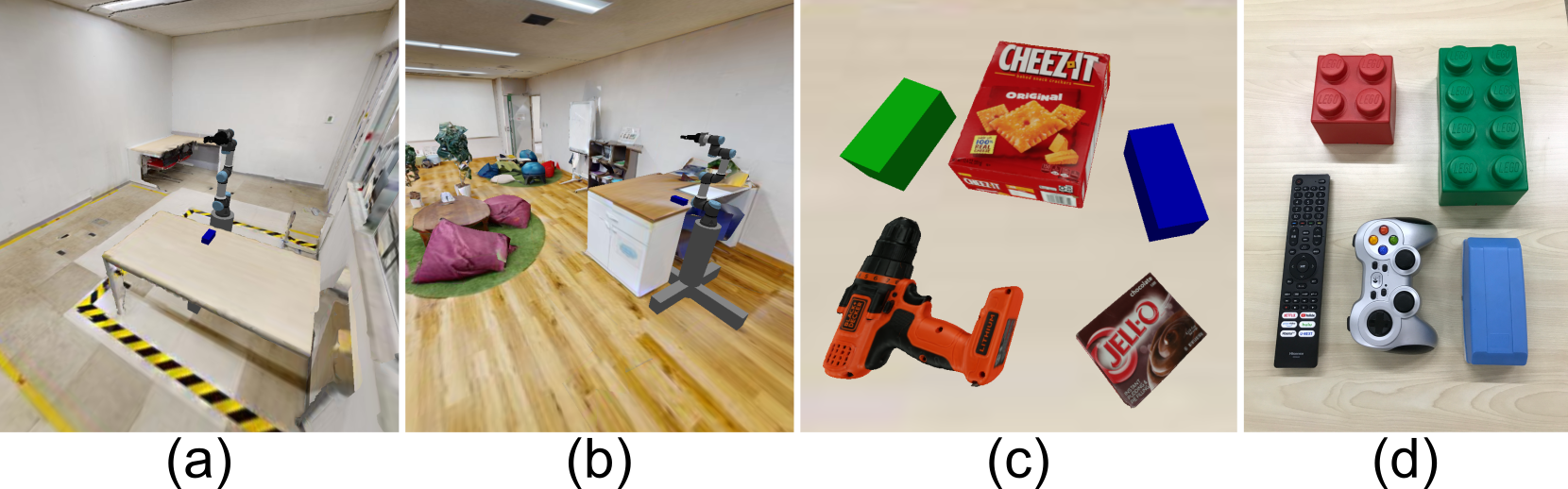}
  \caption{\small {\bf Simulator environments and objects for evaluation.}
  (a) (b) Simulator environments.
  (c) Objects in the simulator.
  (d) Objects for real robot experiments.
  }
  \label{fig:env-obj}
  \vspace{-5mm}
\end{figure}

We evaluated the proposed control policy and system quantitatively for two tasks.
The first is a random pose reaching, which evaluates the control ability from and to a variety of arm poses.
The second is a more realistic task for a robotic manipulator, i.e., reaching a randomly placed object on a desk.
If a grasping posture for the object is given as the target image, the robot can grasp the object after reaching it.
Figs.~\ref{fig:env-obj} (a) and (b) show examples of environments for simulation,
which are 3D-scanned office buildings with different arrangements and room sizes.

The proposed system described in Sec.~\ref{sec:ctrl} was run at a control frequency of 2 Hz.
In addition, we evaluated the following baseline methods for comparison purposes.
\subsubsection{Runtime Optimization (RO)\cite{finn2017}}
Following \cite{finn2017}, we used a cross-entropy method stochastic optimization to generate actions in RO.
$M$ sets of $N$ actions were sampled from a Gaussian distribution and $J_i$ for each action was evaluated using VUNet-6DoF.
$K$ sets of actions with the smallest $J_i$ were selected to update the distribution.
The parameters were selected as $M=20$, $K=10$.
The optimization loop was iterated six times to assure a control frequency of 2 Hz.
\subsubsection{Imitation Learning (IL)\cite{codevillaE2E2018}}
The IL policy is a neural network with the same structure as EnvNet/ObjNet; however,
it was trained without VUNet-6DoF to minimize the mean squared error of actions between the output actions and reference actions from the dataset.

\subsection{Quantitative Analysis}
\subsubsection{Random Pose Reaching}
The random pose reaching task was evaluated as follows.
Before evaluation, the camera image and pose data were collected
while the robot was moving toward randomly sampled 6D poses.
It is noteworthy that the simulation data were collected in the same three environments with the Arm data acquisition.
The real robot experiments with UR3e were performed in a real-world environment corresponding to Fig.~\ref{fig:env-obj} (a).

On the tests, the current and target states were sampled from the data sequence and fed into the control system.
The error for each attempt was recorded when the composed error with the target pose took the minimum value.
The composed error was defined as $e_p + 0.1 e_r$, where $e_p$ is the distance in meters and $e_r$ is the rotation angle in radians.
Each attempt was completed in a time limit of 15 s, or when the controller fell into the stuck state.
The success rate was defined as the ratio of trials with $e_p$ less than 5 cm.
Approximately 750 and 100 trials for each condition were recorded in simulated and real environments, respectively.
Only EnvNet was activated in our system for this test in the static environments.

Table~\ref{tab:result-rp} summarizes the scores for each method and dataset.
In the simulation, both IL and MPC trained with the Arm dataset show comparably good scores,
as well as our method of MPC with the GS$^+$ dataset.
As expected, it was difficult for the policies trained with GS$^-$ to control the 6-DoF arms
as they always output actions with zero $z$, $\alpha$, and $\beta$.
The RO method shows a relatively high variation in the results.
While it showed high accuracy when the target state was close to the starting point,
it failed to output optimized actions when the target was far.
This variance may be suppressed by employing larger $M$ and $K$ parameters of the method;
however, that requires more computing power than a standard PC for the system to run at the given rate.

In the real robot experiment, policies with the Arm dataset show deteriorated performances.
Although the Arm dataset contains data collected in a simulated room where the real experiment was conducted,
the policies trained with this dataset could not work well in the real environment, i.e., they failed in sim-to-real transfer.
Conversely, performances of IL and MPC with GS$^+$ dataset are as good in real-world experiments as in the simulation.
Our method of MPC with GS$^+$ shows the best overall performance.
The performance of Arm data policies in reality would be increased
if millions of real-world datasets could be collected with robotic arms placed in various environments.
However, our method achieved good performance without such high-cost data collection,
through the data collected by a wheeled robot.

\begin{table}[t!]
  \vspace{2mm}
  \caption{\small \label{tab:result-rp} {\bf Results of Random Pose Reaching.} SR: Success Rate in \%. $e_p$: mean (standard deviation) of position error in cm. $e_r$: mean (standard deviation) of rotation error in radians.}
  \vspace{-2mm}
  \begin{center}
\resizebox{0.99\columnwidth}{!}{
  \setlength\tabcolsep{0.08cm}
  \begin{tabular}{ll|ccc|ccc}
\hline
\multicolumn{2}{l}{Method}  & \multicolumn{3}{|c|}{Sim} & \multicolumn{3}{c}{Real} \\
Training                                 & Data   & SR & $e_p$     & $e_r$       & SR & $e_p$   & $e_r$       \\ \hline
RO {\scriptsize \cite{finn2017}}         & GS$^-$ & 44 & 7.1 (5.4) & 0.32 (0.24) & 19 & 16 (11) & 0.53 (0.41) \\
                                         & Arm    & 68 & 4.9 (6.6) & 0.18 (0.27) & 47 & 11 (12) & 0.41 (0.44) \\
                                         & GS$^+$ & 56 & 5.9 (6.3) & 0.22 (0.27) & 31 & 15 (12) & 0.44 (0.45) \\ \hline
IL {\scriptsize \cite{codevillaE2E2018}} & GS$^-$ & 46 & 6.9 (5.5) & 0.32 (0.23) & 19 & 15 (11) & 0.52 (0.39) \\
                                         & Arm    & 81 & 3.4 (4.4) & 0.09 (0.18) & 19 & 16 (12) & 0.54 (0.41) \\
                                         & GS$^+$ & 43 & 6.9 (5.0) & 0.22 (0.25) & 48 & 10 (11) & 0.35 (0.37) \\ \hline
MPC {\scriptsize \cite{hirose2019deep}}  & GS$^-$ & 47 & 7.1 (6.0) & 0.31 (0.24) & 20 & 14 (10) & 0.53 (0.41) \\
                                         & Arm    & 85 & 4.0 (3.5) & 0.07 (0.17) & 17 & 15 (11) & 0.46 (0.39) \\
\bf{(Ours)}                  & \bf{GS$^+$} &\bf{89} & \bf{3.1 (2.6)} & \bf{0.06 (0.13)} & \bf{90} & \bf{3.5 (6.1)} & \bf{0.11 (0.23)} \\ \hline
\end{tabular}
}
\end{center}
  \vspace{-6mm}
\end{table}
\subsubsection{Object Reaching}
For object reaching, several starting poses were predetermined before the evaluation.
The origin of the object was determined on the desk, and a grasping posture corresponding to the object position was defined as the original target pose, where the target camera image $I_G$ was taken.
At the beginning of each attempt, the robot took one of the starting poses,
and the object was placed at a random position on the desk.
The object position $(x, y)$ was sampled from two uniform distributions for $x$ and $y$, with fixed ranges centered at the origin.
The target pose of the arm was also calculated using object displacement.
The ranges of the uniform distribution for object position sampling were 0.35 m in $x$ and 0.20 m in the $y$ directions in the simulated experiments.
For real experiments, the random positions were discretized on a grid with a spacing of 0.025 m within the range to facilitate manual object placement.
Around 1,500 and 100 trials per condition were run in three simulated environments and a real environment with five objects each,
as shown in Figs.~\ref{fig:env-obj} (c) and (d), respectively.
The thresholds for the selector were set $\eta_p=0.22$ m and $\eta_r=0.17$ rad.

Table~\ref{tab:result-obj} lists the scores of our methods.
While ``Ours (EnvNet)'' is the system with EnvNet only,
two policies are activated for ``Ours (Full)''.
The results clearly show that our full system with EnvNet and ObjNet outperforms the single policy case in both the simulation and real robot experiments.
The EnvNet-only system almost ignored displaced objects and tended to take a fixed pose at the origin of target-acquisition.
The $e_r$ for the full system is slightly larger than the EnvNet-only system
because our combined score $e_p + 0.1 e_r$ for evaluation mildly prioritizes $e_p$ over $e_r$.

\begin{table}[t!]
  \vspace{2mm}
  \caption{\small \label{tab:result-obj} {\bf Result of Object Reaching.}}
  \vspace{-3mm}
  \begin{center}
\resizebox{0.99\columnwidth}{!}{
  \setlength\tabcolsep{0.08cm}
    \begin{tabular}{l|ccc|ccc}
    \hline
    \multicolumn{1}{c}{} & \multicolumn{3}{|c|}{Sim} & \multicolumn{3}{c}{Real} \\
Method        & SR       & $e_p$           & $e_r$            & SR      & $e_p$          & $e_r$            \\ \hline
Ours (EnvNet) & 36       & 7.6 (5.0)       & \bf{0.11 (0.11)} & 54      & 5.6 (4.4)      & \bf{0.11 (0.17)} \\
Ours (Full)   & \bf{65}  & \bf{5.9 (6.4)}  & 0.15 (0.12)      & \bf{90} & \bf{3.6 (5.0)} & 0.14 (0.27)      \\ \hline
    \end{tabular}
}
  \end{center}
  \vspace{-3mm}
\end{table}

\subsection{Qualitative Analysis}
\begin{figure}
  \centering
  \includegraphics[width=8.5cm]{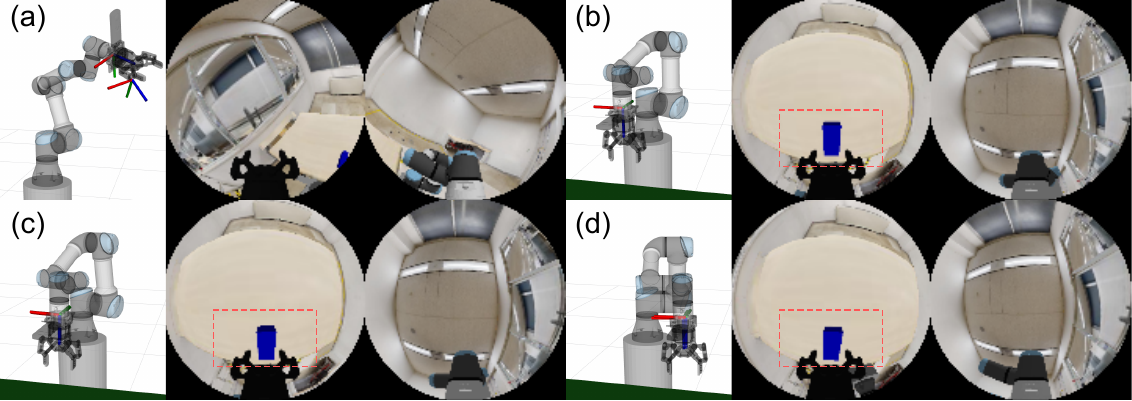}
  \caption{\small {\bf Visualization of object reaching operation.}
  (a) Starting state. (b) Target state. (c) Final state.
  (d) Another final state with an object displacement by $(x, y) = (-0.211, -0.014)$ meters.
  Left is robot model visualization and right is dual-fisheye camera image.
  Dotted rectangles in camera images indicate RoIs of ObjNet.
  }
  %The end-effector pose is visualized as orthogonal axes, and another axes near the end-effector indicates the action $\mathbf{a}(0)$ as relative pose.
  \label{fig:visual}
  \vspace{-4mm}
\end{figure}

Finally, an example of the object reaching in simulation is visualized in Fig.~\ref{fig:visual}.
Our system was given a target image to hold the hands in front of a box on the desk, as shown in Fig.~\ref{fig:visual} (b).
Starting from the state Fig.~\ref{fig:visual} (a), EnvNet navigated the end-effector toward the final state Fig.~\ref{fig:visual} (c).
Fig.~\ref{fig:visual} (d) shows the final state when the object was placed at a different position from the position of target-image acquisition.
In this case, ObjNet was activated to guide the hand to the object.

\section{Conclusions}
In this paper, we showed that a DNN-based control policy for a 6-DoF robot can be trained
with a dataset collected only by a 3-DoF wheeled robot traveling on flat floors
using data augmentation leveraging omnidirectional vision.
The control system utilized another local control policy for object reaching,
which is trained to focus on a limited RoI rather than the entire FoV of the omnidirectional image.

There are remaining future works toward the achievement of more realistic manipulation tasks.
For instance, grasping posture is known in advance and given as the target image for our system.
Thus, a grasping posture estimation technique will be required to grasp unknown objects.
Our threshold-based selection algorithm of global/local policies is not optimal
because it imposes limits on the applicable range of the local policy.
However, the proposed concept of trainable DoF expansion could be applied to a broad variety of robots such as legged and aerial robots.

%\addtolength{\textheight}{-12cm}

%\section*{Appendix}

%\section*{Acknowledgment}
%Thanks

\clearpage

\bibliographystyle{IEEEtran}
\vskip-\parskip
\begingroup
%\footnotesize
\bibliography{egbib}
\endgroup

\end{document}